\documentclass{article}
\usepackage[utf8]{inputenc}

\usepackage{times}

\usepackage[]{url}
\usepackage{soul}

\usepackage{amsmath,amsfonts,amssymb}
\usepackage{mathabx}
\newtheorem{definition}{Definition}

\usepackage{tikz}
\usetikzlibrary{arrows,automata,positioning}
\usetikzlibrary{arrows.meta}

\usepackage{dialogue}

\usepackage{tabu}
\usepackage{booktabs}
\usepackage{array}
\newcommand{\ie}{i.e.}

\newcommand{\tuple}[1]{\ensuremath{\langle #1 \rangle}}
\newcommand{\set}[1]{\ensuremath{\{#1\}}}

\newcommand{\argument}[1]{\ensuremath{\textrm{\textbf{#1}}}}
\newcommand{\arga}{\argument{a}}
\newcommand{\argb}{\argument{b}}
\newcommand{\argc}{\argument{c}}
\newcommand{\argd}{\argument{d}}

\newcommand{\textlabel}[1]{\ensuremath{\mathrm{\mathtt{#1}}}}
\newcommand{\textin}{\ensuremath{\textlabel{in}}}
\newcommand{\textout}{\ensuremath{\textlabel{out}}}
\newcommand{\textundec}{\ensuremath{\textlabel{undec}}}
\newcommand{\Labfun}{\ensuremath{\mathcal{L}ab}}
\newcommand{\setoflabellings}{\ensuremath{\mathfrak{L}}}

\newcommand{\AFname}{\ensuremath{AF}}
\newcommand{\setargs}{\ensuremath{\mathcal{A}}}
\newcommand{\setattacks}{\ensuremath{\mathcal{R}}}
\newcommand{\AF}[1]{\tuple{\setargs_{#1}, \setattacks_{#1}}}
\newcommand{\anAF}{\AF{}}
\newcommand{\AFsymbol}[1]{\ensuremath{\Gamma_{#1}}}
\newcommand{\anAFsymbol}{\ensuremath{\Gamma}}
\newcommand{\attacks}[2]{\ensuremath{#1 \rightarrow #2}}

\newcommand{\dungconffree}{conflict--free}
\newcommand{\dungacceptable}{acceptable}

\newcommand{\dungadmissible}{admissible}

\newcommand{\dungpreferred}{preferred}

\newcommand{\gensem}{\sigma}
\newcommand{\setgenext}[2]{\mathcal{E}_{#1}(#2)}

\newcommand{\PR}{\ensuremath{\mathsf{PR}}}

\newcommand{\charfun}[1]{\ensuremath{\mathcal{F}_{#1}}}

\newcommand{\attackers}[1]{#1^{-}}
\newcommand{\attacked}[1]{#1^{+}}

\newcommand{\aset}{\ensuremath{S}}

\newcommand{\setnodes}{\ensuremath{\mathcal{N}}}
\newcommand{\setinodes}{\ensuremath{\setnodes_I}}
\newcommand{\setsnodes}{\ensuremath{\setnodes_S}}
\newcommand{\setschemes}{\ensuremath{\mathcal{S}}}
\newcommand{\setrules}{\ensuremath{\setschemes^{R}}}
\newcommand{\setconflicts}{\ensuremath{\setschemes^{C}}}
\newcommand{\setpreferences}{\ensuremath{\setschemes^{P}}}
\newcommand{\setranodes}{\ensuremath{\setsnodes^{RA}}}
\newcommand{\setpanodes}{\ensuremath{\setsnodes^{PA}}}
\newcommand{\setcanodes}{\ensuremath{\setsnodes^{CA}}}
\newcommand{\fulfils}{\ensuremath{\mathsf{fulfils}}}

\usepackage{authblk}

\title{On Natural Language Generation of Formal Argumentation}

\author[1]{Federico Cerutti}
\author[2]{Alice Toniolo}
\author[3]{Timothy J. Norman}
\affil[1]{Cardiff University\\
School of Computer Science \& Informatics\\
Cardiff, UK}
\affil[2]{University of St.\ Andrews\\
School of Computer Science\\
St.\ Andrews, UK}
\affil[3]{University of Southampton\\
Department of Electronics and Computer Science\\
Southampton, UK}

\date{}

\begin{document}

\maketitle

\begin{abstract}
In this paper we provide a first analysis of the research questions that arise when dealing with the problem of communicating pieces of formal argumentation through natural language interfaces.
It is a generally held opinion that formal models of argumentation naturally capture human argument, and some preliminary studies have focused on justifying this view. Unfortunately, the results are not only inconclusive, but seem to suggest that explaining formal argumentation to humans is a rather articulated task.
Graphical models for expressing argumentation-based reasoning are appealing, but often humans require significant training to use these tools effectively. 
We claim that natural language interfaces to formal argumentation systems offer a real alternative, and may be the way forward for systems that capture human argument.
\end{abstract}

\section{Introduction}

Our aim here is to explore the challenges that we need to face when thinking about natural language interfaces to formal argumentation.
Dung \cite{dung1995} states that formal argumentation ``captures naturally the way humans argue to justify their solutions to many social problems.'' This is one of the most common claims used to support research in formal argumentation. More recently there have been a number of empirical studies to investigate this claim \cite{Rahwan2009,Cerutti2014f,RosenfeldKraus16,RosenfeldKraus16bis}. The results, however, have been far from conclusive.

The use of graphical models to represent arguments is the most common approach used in the formal argumentation community to capture argument structures. This has been successfully applied in a number of real world situations: Laronge \cite{Laronge2012} (an American barrister and researcher), for example, describes how he used argumentation graphs during trials. Despite this, our claim is that to produce and to consume a graphical representation of a structure of arguments there is need for significant levels of training.

Instead of training users on another (graphical) language for representing argument structures, we can leverage our societal model, through which we are trained in reading and writing; that is, using natural language. We claim that natural language representations of formal argumentation are the way forward to develop formal models that capture human argumentation. In this paper we investigate one aspect of natural language interfaces to formal argumentation: moving from formal arguments to natural language text by exploiting Natural Language Generation (NLG) systems. We ground our investigation on an existing example (Section \ref{sec:running-example}) from collaboration between the BBC and the Dundee argumentation group \cite{Lawrence2012}, namely an excerpt of the BBC Radio 4 Moral Maze programme from 22nd June 2012. In this way the reader can always relate to the original piece of text from where our investigation started.
The excerpt has been already formalised into an argument network (i.e.\ a graph linking together different pieces of information together to display the web of arguments exchanged). In Figure \ref{fig:dundee1724} and in Section \ref{sec:background} we review all the necessary elements  for our investigation: the notion of argumentation schemes; how to represent argument networks in the Argument Interchange Format (AIF); a (simple) approach to structured argumentation to build arguments and approximate arguments from an AIF argument network; Dung's theory of argumentation; and basic elements of NLG.

In NLG one of the most difficult tasks is to determine the communicative goal; i.e. deciding what we would like to communicate. Therefore, Section \ref{sec:nlgarg} is entirely dedicated to investigating relevant communicative goals in the context of formal argumentation.

The result we wish to achieve in this paper is a blueprint that outlines the complex research questions and their dense interconnections that our community needs to address in order to identify models that naturally capture human argumentation.


\section{Running Example}
\label{sec:running-example}

On 22nd June 2012, in the middle of the European debt crisis, the Moral Maze program on BBC Radio 4 addressed the topic of individual and national debt. Among others, Nick Dearden, director of the Jubilee Debt Campaign, and Claire Fox, from the Institute of Ideas, were `witnesses' (contributors offering a specific point of view in the debate) during the program.

\begin{figure}
    \centering
    \includegraphics[width=\linewidth]{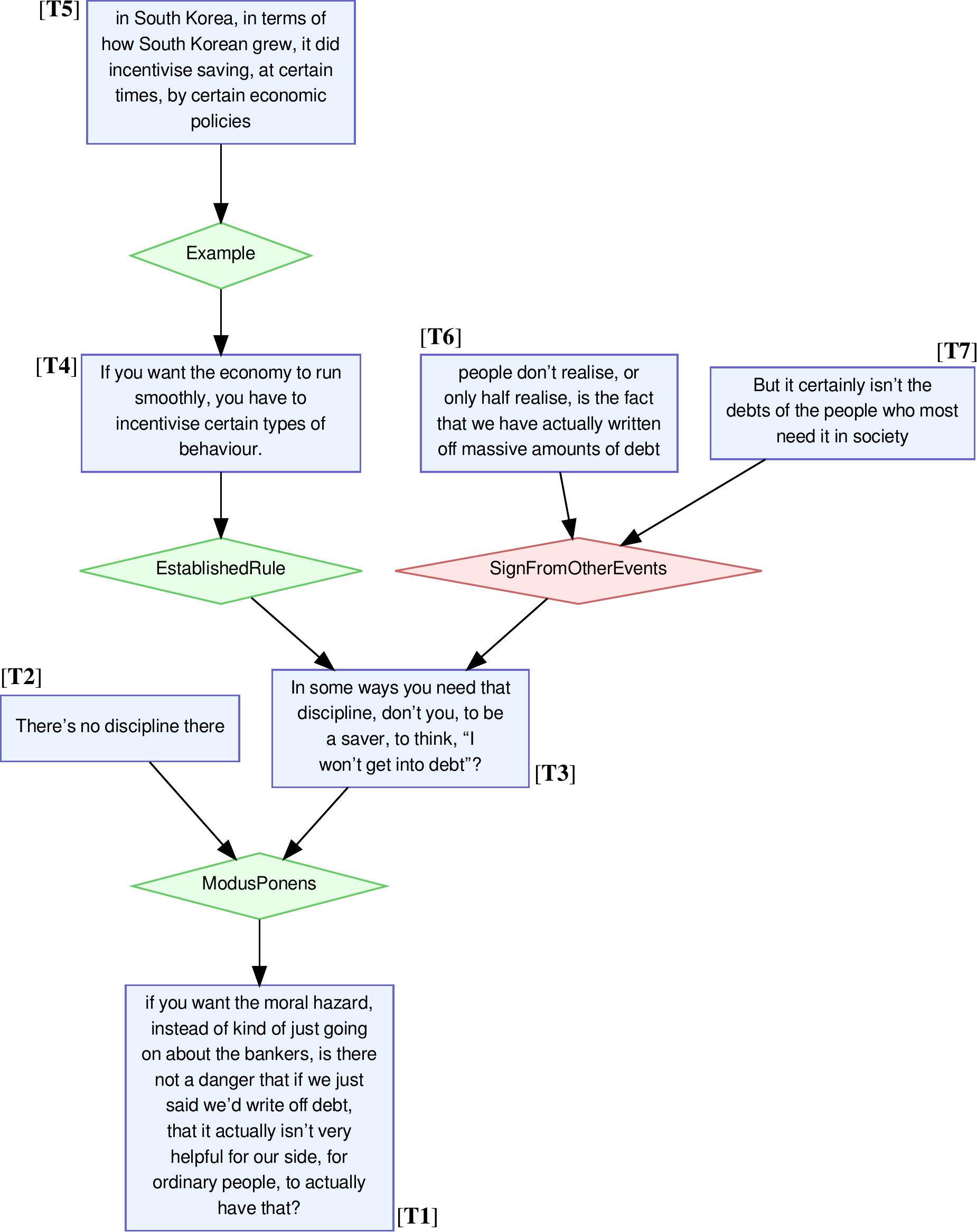}
    \caption{Argument network for the running example, as obtained from \protect\url{http://aifdb.org/argview/1724} (edited version)}
    \label{fig:dundee1724}
\end{figure}

What follows is an excerpt that has been analysed to identify the argument network in the dialogue and made available at \url{http://aifdb.org/argview/1724} \cite{Lawrence2012}. In this paper we focus on the sub-part of the argument network depicted in Figure \ref{fig:dundee1724} and we highlight in the text the elements that contribute to the generation of such an argument network.

\begin{dialogue}
\speak{Claire Fox} I understand that. I suppose my concern is just this: I want the freedom to be able to write off debts but --- I'm sure you recognise this --- there is this sort of sense amongst a lot of young people, who just think, ``I want that, so I'll have that now. Thank you.'' And so, \ul{if you want the moral hazard, instead of kind of just going on about the bankers, is there not a danger that if we just said we'd write off debt, that it actually isn't very helpful for our side, for ordinary people, to actually have that?} [\textbf{T1}] \ul{There's no discipline there} [\textbf{T2}]. \ul{In some ways you need that discipline, don't you, to be a saver, to think, ``I won't get into debt?''} [\textbf{T3}]

\speak{Nick Dearden} In some ways I agree with you. \ul{If you want the economy to run smoothly, you have to incentivise certain types of behaviour.} [\textbf{T4}] So, for example \ul{in South Korea, in terms of how South Korean grew, it did incentivise saving, at certain times, by certain economic policies} [\textbf{T5}]. On the other hand, I think what \ul{people don't realise, or only half realise, is the fact that we have actually written off massive amounts of debt} [\textbf{T6}]. \ul{But it certainly isn't the debts of the people who most need it in society} [\textbf{T7}]. 
\end{dialogue}

\section{Background}
\label{sec:background}

\subsection{Argumentation Schemes}


Argumentation schemes \cite{waltonetal2008,Walton2009} are abstract reasoning patterns commonly used in everyday conversational argumentation, legal, scientific argumentation, etc. Schemes have been derived from empirical studies of human argument and debate. They can capture traditional deductive and inductive approaches as well as plausible reasoning. Each scheme has a set of critical questions that represents standard ways of critically probing into an argument to find aspects of it that are open to criticism.

For instance, in the dialogue reported in Section \ref{sec:running-example}, part of Nick's position can be mapped into an argument from example that has the following structure:

\noindent
\begin{tabbing}
Conclusion: \= \kill
Premise: \> In this particular case, individual \emph{a} has property \emph{F} and also property \emph{G}.\\
Conclusion: \> Therefore, generally, if \emph{x} has property \emph{F}, the it also has property \emph{G}.
\end{tabbing}

One of the critical questions here is: ``Is the proposition claimed in the premise in fact true?''
 


\subsection{Representing an Argument Network}

The Argument Interchange Format (AIF) \cite{CHESNEVAR2006,Rahwan2009} is the current proposal for a standard notation for argument structures. It is based on a graph that specifies two types of nodes: information nodes (or I-nodes) and scheme
nodes (or S-nodes). These are represented by two disjoint sets, $\setinodes \cup \setsnodes = \setnodes$ and $\setinodes \cap \setsnodes = \emptyset$, where information nodes represent claims, premises, data, etc., and scheme nodes capture the application of patterns of reasoning belonging to a set $\setschemes = \setrules \cup \setconflicts \cup \setpreferences$, $\setrules \cap \setconflicts = \setconflicts \cap \setpreferences = \setpreferences \cap \setrules = \emptyset$. Reasoning patterns can be of three types: rule of inference $\setrules$; criteria of preference $\setpreferences$; and criteria of conflicts $\setconflicts$.

The relation $\fulfils \subseteq \setsnodes \times \setschemes$ expresses that a scheme node instantiates a particular scheme. Scheme nodes, moreover, can be one of three types: rule of inference application nodes $\setranodes$; preference application nodes $\setpanodes$; or conflict application nodes $\setcanodes$, with $\setschemes = \setranodes \cup \setpanodes \cup \setcanodes$, and $\setranodes \cap \setpanodes = \setpanodes \cap \setcanodes = \setcanodes \cap \setranodes = \emptyset$.

Nodes are connected by edges whose semantics is implicitly defined by their use. For instance, an information node connected to a RA scheme node, with the arrow terminating in the latter, would suggest that the information node serves as a premise for an inference rule. Figure \ref{fig:dundee1724} shows an AIF representation of the arguments exchanged in the dialogue introduced in Section \ref{sec:running-example}. Rectangular nodes represent information nodes, while rhombic ones represent scheme nodes: green for RA nodes, and red for CA nodes.

\subsection{Deductive Argumentation}
\label{sec:simplelogic}

Using deductive argumentation means that each argument is defined using a logic, and in the following we adopt the simple, but elegant logic proposed by Besnard \& Hunter \cite{Besnard2014a}. Thus, we let $\mathcal{L}$ be a logical language.
If $\alpha$ is an atom in $\mathcal{L}$, then $\alpha$ is a \emph{positive literal} in $\mathcal{L}$ , and $\lnot \alpha$ is a \emph{negative literal} in $\mathcal{L}$. For a literal $\beta$, the \emph{complement} of the positive literal $\beta = \alpha$ is $\overline{\beta} = \lnot \alpha$ (resp. if $\beta = \lnot \alpha$ is not a positive literal, its complement is the positive literal $\overline{\beta} = \alpha$).


A \emph{simple rule} is of the form $\alpha_1 \land \ldots \land \alpha_k \rightarrow \beta$ where $\alpha_1, \ldots, \alpha_k, \beta$ are literals. A \emph{simple logical knowledge base} is a set of literals and a set of simple rules. Given a simple logic knowledge base, $\Delta$, the \emph{simple consequence relation} $\vdash_s$ is defined, such that $\Delta \vdash_s \beta$ if and only if there is a rule $\alpha_1 \land \ldots \land \alpha_n \rightarrow \beta \in \Delta$ and $\forall i$ either $\alpha_i \in \Delta$ or $\Delta \vdash_s \alpha_i$. Now, given $\Phi \subseteq \Delta$ and a literal $\alpha$, $\tuple{\Delta, \alpha}$ is a \emph{simple argument} if and only if $\Phi \vdash_s \alpha$ and $\nexists \Phi' \subsetneq \Phi$ such that $\Phi' \vdash_s \alpha$.
$\Phi$ is the support (or premises, assumptions) of the argument, and $\alpha$ is the claim (or conclusion) of the argument.
Given an argument $\arga = \tuple{\Phi, \alpha}$, the function $\mathsf{Support}(\arga)$ returns $\Phi$, and $\mathsf{Claim}(\arga)$ returns $\alpha$.

For simple arguments \arga\ and \argb\ we consider the following types of simple attack:
\begin{itemize}
    \item \arga\ is a \emph{simple undercut} of \argb\ if there is a simple rule $\alpha_1 \land \alpha_k \rightarrow \beta$ in $\mathsf{Support}(\argb)$ and there is an $\alpha_i \in \set{\alpha_1, \ldots, \alpha_k}$ such that $\mathsf{Claim}(\arga)$ is the complement of $\alpha_i$;
    \item \arga\ is a \emph{simple rebut} of \argb\ if $\mathsf{Claim}(\arga)$ is the complement of $\mathsf{Claim}(\argb)$.
\end{itemize}

Following Black \& Hunter \cite{Black2012}, an \emph{approximate argument} is a pair $\tuple{\Phi, \alpha}$. If $\Phi \vdash_s \alpha$, then $\tuple{\Phi, \alpha}$ is also \emph{valid}; if  $\Phi \nvdash_s \bot$, then $\tuple{\Phi, \alpha}$ is also \emph{consistent}; if $\Phi \vdash_s \alpha$, and there is no $\Phi' \subsetneq \Phi$ such that $\Phi' \vdash_s \alpha$, then $\tuple{\Phi, \alpha}$ is also \emph{minimal}; if $\Phi \vdash_s \alpha$, and $\Phi \nvdash_s \bot$, then $\tuple{\Phi, \alpha}$ is also \emph{expansive} (i.e.\ it is valid and consistent, but it may have unnecessary premises).

Building on top of Figure \ref{fig:dundee1724} and transforming each $\setranodes$ node into a simple rule, a simple knowledge base for our running example is:
\[\Delta_m = \left\{
\begin{array}{l}
     \mbox{[\textbf{T1}]}, \mbox{[\textbf{T2}]}, \mbox{[\textbf{T3}]}, \mbox{[\textbf{T4}]}, \mbox{[\textbf{T5}]}, \\
     \mbox{[\textbf{T5}]} \rightarrow \mbox{[\textbf{T4}]}, \\
     \mbox{[\textbf{T4}]} \rightarrow \mbox{[\textbf{T3}]}, \\
     \mbox{[\textbf{T2}]} \land \mbox{[\textbf{T3}]} \rightarrow \mbox{[\textbf{T1}]}, \\
     \mbox{[\textbf{T6}]} \land \mbox{[\textbf{T7}]} \rightarrow \overline{\mbox{[\textbf{T3}]}}
\end{array}
\right\}
\]

Therefore, the following are the simple arguments that can be built from $\Delta_m$:

\[\setargs_m = \left\{
\begin{array}{l}
     \arga = \tuple{\set{\mbox{[\textbf{T5}]}, \mbox{[\textbf{T5}]} \rightarrow \mbox{[\textbf{T4}]}}, \mbox{[\textbf{T4}]}},\\
     \argb = \tuple{\set{\mbox{[\textbf{T4}]}, \mbox{[\textbf{T4}]} \rightarrow \mbox{[\textbf{T3}]}} \mbox{[\textbf{T3}]}},\\
     \argc = \tuple{\set{\mbox{[\textbf{T2}]}, \mbox{[\textbf{T3}]}, \mbox{[\textbf{T2}]} \land \mbox{[\textbf{T3}]} \rightarrow \mbox{[\textbf{T1}]}}, \mbox{[\textbf{T1}]}},\\
     \argd = \tuple{\set{\mbox{[\textbf{T6}]}, \mbox{[\textbf{T7}]}, \mbox{[\textbf{T6}]} \land \mbox{[\textbf{T7}]} \rightarrow \overline{\mbox{[\textbf{T3}]}}}, \overline{\mbox{[\textbf{T3}]}}} \\
\end{array}
\right\}
\]

\noindent
with \argd\ rebutting \argb, and \argd\ undercutting \argc.

However, there are many more approximate arguments. For instance, $\argc' = \linebreak \tuple{\set{\mbox{[\textbf{T5}]}, \mbox{[\textbf{T4}]}, \mbox{[\textbf{T3}]}, \mbox{[\textbf{T2}]}, \mbox{[\textbf{T5}]} \rightarrow \mbox{[\textbf{T4}]}, \mbox{[\textbf{T4}]} \rightarrow \mbox{[\textbf{T3}]}, \mbox{[\textbf{T2}]} \land \mbox{[\textbf{T3}]} \rightarrow \mbox{[\textbf{T1}]}}, \mbox{[\textbf{T1}]}}$ is an approximate argument in favour of \mbox{[\textbf{T1}]} taking into consideration all the inferences that might help concluding it. Conversely, $\argc'' = \tuple{\set{}, \mbox{[\textbf{T1}]}}$ is the minimal (invalid) approximate argument in favour of \mbox{[\textbf{T1}]}.

\subsection{Abstract Argumentation}

An argumentation framework \cite{dung1995} consists of a set of arguments 
and a binary attack relation between them. 

\begin{definition}
An \emph{argumentation framework} (\AFname) is a pair $\anAFsymbol = \anAF$ 
where $\setargs$ is a set of arguments and $\setattacks \subseteq \setargs \times \setargs$. 
We say that \argb{} \emph{attacks} \arga{} iff $\tuple{\argb,\arga} \in \setattacks$, also denoted as $\attacks{\argb}{\arga}$.
The set of attackers of an argument $\arga$ will be denoted as 
$\attackers{\arga} \triangleq \set{\argb : \attacks{\argb}{\arga}}$,
the set of arguments attacked by $\arga$ will be denoted as
$\attacked{\arga} \triangleq \set{\argb : \attacks{\arga}{\argb}}$.
We also extend these notations to sets of arguments, \ie{} given $E, S \subseteq \setargs$, $\attacks{E}{\arga}$ iff $\exists \argb \in E$ s.t. $\attacks{\argb}{\arga}$; 
$\attacks{\arga}{E}$ iff $\exists \argb \in E$ s.t. $\attacks{\arga}{\argb}$; 
$\attacks{E}{S}$ iff $\exists \argb \in E, \arga \in S$ s.t. $\attacks{\argb}{\arga}$; 
$\attackers{E} \triangleq \set{\argb \mid \exists \arga \in E, \attacks{\argb}{\arga}}$
and
$\attacked{E} \triangleq \set{\argb \mid \exists \arga \in E, \attacks{\arga}{\argb}}$.
\end{definition}

Each argumentation framework, therefore, has an associated directed graph where the vertices are the arguments, and the edges are the attacks.

The basic properties of \dungconffree ness, acceptability, and admissibility of a set of arguments 
are fundamental for the definition of argumentation semantics.

\begin{definition}
\label{def:recall}
Given an $\AFname$ $\anAFsymbol = \anAF$:
\begin{itemize}

  \item a set $\aset \subseteq \setargs$ is a \emph{\dungconffree} set of $\anAFsymbol$
        if $\nexists~ \arga, \argb \in \aset$ s.t. $\attacks{\arga}{\argb}$;
  \item an argument $\arga \in \setargs$ is \emph{\dungacceptable} with respect to a set $\aset \subseteq \setargs$ of  $\anAFsymbol$
        if $\forall \argb \in \setargs$ s.t. $\attacks{\argb}{\arga}$, 
        $\exists~ \argc \in \aset$ s.t. $\attacks{\argc}{\argb}$;


  \item a set $\aset \subseteq \setargs$ is an \emph{\dungadmissible} set of $\anAFsymbol$
        if $\aset$ is a \dungconffree{} set of $\anAFsymbol$ and every element of $\aset$ is \dungacceptable{} with respect to $\aset$,
        \ie{} $\aset \subseteq \charfun{\anAFsymbol}(\aset)$.

\end{itemize}
\end{definition}

An argumentation semantics $\gensem$ prescribes for any \AFname{} $\anAFsymbol$ a set of \emph{extensions}, denoted as $\setgenext{\gensem}{\anAFsymbol}$, namely a set of sets of arguments satisfying the conditions dictated by $\gensem$. 
For instance, here is the definition of  preferred (denoted as $\PR$) semantics.

\begin{definition}
\label{def_sem_recall}
Given an $\AFname$ $\anAFsymbol = \anAF$, a set $\aset \subseteq \setargs$ is a \emph{\dungpreferred{} extension} of $\anAFsymbol$, 
             \ie{} $\aset \in \setgenext{\PR}{\anAFsymbol}$, 
              iff $\aset$ is a maximal (w.r.t. set inclusion) \dungadmissible{} set of  $\anAFsymbol$.
\end{definition}

Given a semantics $\sigma$, an argument \arga\ is said to be \emph{credulously accepted} w.r.t. $\sigma$ if \arga\ belongs to at least one $\sigma$-extension. \arga\ is \emph{skeptically accepted} w.r.t. $\sigma$ if $\arga$ belongs to all the $\sigma$-extensions.

It can be noted that each complete extension $\aset$ implicitly defines a three-valued \emph{labelling function} \Labfun{} on the set of arguments: 
an argument $\arga$ is labelled \textin{} iff $\arga \in \aset$;
is labelled \textout{} iff $\exists~ \argb \in \aset$ s.t. $\attacks{\argb}{\arga}$;
and is labelled \textundec{} if neither of the above conditions holds. 
In the light of this correspondence, argumentation semantics can be equivalently defined 
in terms of labellings rather than of extensions \cite{Caminada2006,KER2011}.

We can now introduce the concept of the \emph{issues} of an argumentation framework 
whose status is enough to determine the status of all the
arguments in the framework \cite{gabbay2009b,Booth2012,Booth2014}.

\begin{definition}
\label{def:issues}
Given an $\AFname$ $\anAFsymbol = \anAF$ and $\setoflabellings$ the set of all complete labellings of $\anAFsymbol$, for any $\arga, \argb \in \setargs$, $\arga \equiv \argb$ iff $\forall \Labfun \in \setoflabellings$, $\Labfun(\arga) = \Labfun(\argb)$; or $\forall \Labfun \in \setoflabellings$, ($\Labfun(\arga) = \textin \iff \Labfun(\argb) = \textout) \land (\Labfun(\arga) = \textout \iff \Labfun(\argb) = \textin$).

The set of arguments in the equivalent class $\equiv$ is the set of \emph{issues} of \anAFsymbol:
\[
    \setgenext{\mbox{\rm issues}}{\anAFsymbol} = \left\{
        \begin{array}{l l}
             \aset \subseteq \setargs ~\mid &  \forall \tuple{\arga, \argb} \in \aset \times \aset, \arga \equiv \argb;\mbox{ and }\\
             & \forall \aset' \supset \aset, \exists \tuple{\argc, \argd} \in \aset' \times \aset', \lnot (\argc \equiv \argd)
        \end{array}
        \right\}
\]
\end{definition}

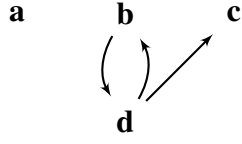
\begin{figure}[t]
  \centering
      \begin{tikzpicture}[->,>=stealth',shorten >=1pt,auto,node
      distance=1.2cm, thick,main node/.style={draw=none,fill=none},scale=1.2, every node/.style={scale=1.2}]
      
      \node[main node] (1) {$\arga$};
      \node[main node] (2) [right of=1] {$\argb$};
      \node[main node] (3) [below of=2] {$\argd$};
      \node[main node] (4) [right of=2] {$\argc$};

      \tikzset{edge/.style = {->,> = latex'}} 
      
      \draw[edge] (2) to[bend right] (3);
      \draw[edge] (3) to[bend right] (2);
      \draw[edge] (3) to (4);
    \end{tikzpicture}

  \caption{The \AFname{} $\AFsymbol{m}$ for Figure \ref{fig:dundee1724} interpreted using deductive argumentation.}
  \label{fig:ex}
\end{figure}

Continuing with our running example, Figure \ref{fig:ex} depicts the argumentation framework from Section \ref{sec:simplelogic} applying deductive argumentation on the argument network of Figure \ref{fig:dundee1724}: $\AFsymbol{m} = \AF{m} = \tuple{\set{\arga, \argb, \argc, \argd}, \set{\attacks{\argd}{\argb}, \attacks{\argb}{\argd}, \attacks{\argd}{\argc}}}$.

There are two preferred extensions: $\setgenext{\PR}{\AFsymbol{m}} = \set{\set{\arga, \argb, \argc}, \set{\arga, \argd}}$.
Moreover, $\set{\argb, \argd} \in \setgenext{\mbox{issues}}{\AFsymbol{m}}$, and if $\Labfun(\argb) = \textin$, $\Labfun(\argd) = \textout$ (and viceversa).

\subsection{Natural Language Generation}
\label{sec:backgroundnlg}

A Natural Language Generation (NLG) system requires \cite{Reiter2000}:
\begin{itemize}
    \item A knowledge source to be used;
    \item A communicative goal to be achieved;
    \item A user model; and
    \item A discourse history.
\end{itemize}

In general, the knowledge source is the information about the domain, while the communicative goal describes the purpose of the text to be generated. The user model is a characterisation of the intended audience, and the discourse history is a model of what has been said so far.

\begin{table}[t]
    \centering
        \caption{Modules and tasks of a NLG system, adapted from \cite[Figure 3.1]{Reiter2000}.}
    
    \begin{tabu}{X[2,l] X[3,l] X[2,l]}
        \toprule
        \textbf{Module} & \textbf{Content task} & \textbf{Structure task}\\
        \midrule
         Document planning & Content determination & Document structuring\\
         Microplanning & Lexicalisation\newline{} Referring expression generation & Aggregation\\
         Realisation & Linguistic realisation & Structure realisation\\
         \bottomrule
    \end{tabu}
    \label{tab:nlgmodules}
\end{table}

An NLG system divides processing into a pipeline \cite{Reiter2000} composed of the three stages described in Table \ref{tab:nlgmodules}. First it determines the content and structure of a document (\emph{document planning}); then it looks at syntactic structures and choices of words that needs to be used to communicate the content chosen in the document planning (\emph{microplanning}). Finally, it maps the output of the microplanner into text (\emph{realisation}).

Each stage includes tasks that can be primary concerned with either content or structure. Document planning requires content determination --- deciding what information should be communicated in the output document --- and document structuring --- how to order the information to be conveyed.

Microplanning requires (1) lexicalisation --- deciding what syntactic constructions our NLG system should use; (2) referring expression generation --- how to relate with entities; and (3) aggregation --- how to map the structures created by the document planning onto linguistic structures such as sentences and paragraphs.

Document planning and microplanning are the most strategic and complex modules in this pipeline. They focus on identifying the communicative goal and how it relates to the user model, thus producing the blueprint of the document that will be generated. 

It is the responsibility of the document planning module, in particular of the document structuring task, to consider the rhetorical relations (or discourse relations) that hold between messages or groups of messages. For instance, Rhetorical Structure Theory (RST) \cite{Mann1988} stresses the idea that the coherency of a text depends on the relationships between pairs of  \emph{text spans} (an uninterrupted linear interval of text) a \emph{nucleus} and a \emph{satellite}. In \cite{Mann1988} a variety of relationships are provided, but for the purpose of this work, we focus on:

\begin{description}
\item[Evidence] The nucleus contains a claim, while the satellite(s) contain(s) evidence supporting such a claim.

\item[Justify] The nucleus contains a claim, while the satellite(s) contain(s) justification for such a claim.

\item[Antithesis] Nucleus and satellite are in contrast.
\end{description}

Finally, once the more strategical tasks are performed, there is need for linguistic realisation --- from abstract representations of sentences to actual text --- and structure realisation --- converting abstract structures such as paragraphs and sections into the mark-up symbols chosen for the document. The realisation module is the most algorithmic in this pipeline, and there are already several implementations for supporting it, for instance \textsf{SimpleNLG} \cite{gatt2009simplenlg}.

In the next section we highlight the cases of document planning and microplanning we believe are most interesting from an argumentation perspective.

\section{Generating Natural Language Interfaces to Formal Argumentation}
\label{sec:nlgarg}


Let us now return to our running example to illustrate four relevant communicative goals. For the moment we are not making any assumptions about the user model, and we will assume no pre-existing discourse history. Therefore, given that the knowledge source is fixed, we envisage the following four communicative goals, each of which raises interesting and challenging questions:
\begin{enumerate}
    \item Presenting a single argument or an approximate argument;
    \item Presenting  an entire argument network;
    \item Explaining the acceptability status of a single argument or an approximate argument; and
    \item Explaining the extensions, given some semantics.
\end{enumerate}

In the following, we discuss elements of content determination for each of these goals; i.e. deciding what messages should be included in the document to be generated. Examples will be provided based on our running example. In parts, the generated texts will sound a little awkward because we deliberately chose not to modify the content of the arguments. We will elaborate on this in Section \ref{sec:conclusion}. 

\subsection{Presenting a Simple or an Approximate Argument}

Simple  and approximate arguments are composed of premises and a claim. There are two main strategies traditionally adopted to represent such a construct: 
\begin{itemize}
    \item \emph{forward writing}, from premises to claim;
    \item \emph{backward writing},\footnote{Often also named \emph{assert-justify}.} from claim to premises.
\end{itemize}

Let us consider the argument $\argb = \tuple{\set{\mbox{[\textbf{T4}]}, \mbox{[\textbf{T4}]} \rightarrow \mbox{[\textbf{T3}]}}, \mbox{[\textbf{T3}]}}$. In the case of \emph{forward writing}, we can write something like:

\begin{quotation}
\ul{If you want the economy to run smoothly, you have to incentivise certain type of behaviour.} [\textbf{T4}]
\ul{In some ways you need that discipline, don't you, to be a saver, to think, ``I
won't get into debt''?.} [\textbf{T3}]
\end{quotation}

\noindent
An explicit signal such as \fbox{\sf Therefore} can be used to highlight the \emph{Justify} relation:

\begin{quotation}
\ul{If you want the economy to run smoothly, you have to incentivise certain type of behaviour.} [\textbf{T4}]

\noindent
\fbox{\textsf{Therefore}} 
\ul{In some ways you need that discipline, don't you, to be a saver, to think, ``I won't get into debt''?.} [\textbf{T3}]
\end{quotation}

Similarly in the case of \emph{backward writing}, we may write:

\begin{quotation}
\ul{In some ways you need that discipline, don't you, to be a saver, to think, ``I won't get into debt''?.} [\textbf{T3}]

\noindent
\fbox{\textsf{Indeed}} 
\ul{If you want the economy to run smoothly, you have to incentivise certain type of behaviour.} [\textbf{T4}]
\end{quotation}

More interesting is the case of considering an approximate argument. Assuming that the communicative goal here is not to confuse the reader, it is probably reasonable not to include irrelevant elements in the presentation. Relevance theory may be seen as an attempt to analyse an essential feature of most human communication: the expression and recognition of intentions \cite{Grice1989,Sperber2005}.

For instance, the First or Cognitive Principle of Relevance proposed by Sperber and Wilson \cite{Sperber1995,Sperber2005} states: \emph{Human cognition tends to be geared to the maximisation of relevance}. Therefore, it would be evidence of poor judgment to expand the argument by considering \emph{irrelevant} elements. Unfortunately, defining the concept of \emph{relevance} in formal argumentation is a non-trivial task \cite{VanEemerenF2014,Paglieri2014}. To take a reasonable starting point, let us adapt the definition of relevant evidence from the Rule 401 of the Federal Rules of Evidence\footnote{\url{https://www.law.cornell.edu/rules/fre/rule_401} (on 13/05/2017)} to our context. Therefore, an approximate argument is relevant for an argument if:
\begin{enumerate}
\item Its premises have any tendency to make the argument's conclusion more or less probable than it would be without them; and
\item It provides additional information that might advance the debate.
\end{enumerate}

Given this (loose) definition of relevance, we could argue that the approximate argument $\argb' = \tuple{\set{\mbox{[\textbf{T5}]}, \mbox{[\textbf{T5}]} \rightarrow \mbox{[\textbf{T4}]}, \mbox{[\textbf{T4}]} \rightarrow \mbox{[\textbf{T3}]}}, \mbox{[\textbf{T3}]}}$ is relevant for \argb. Similarly as before, there are different possibilities to write such a chain of inferences. In addition, it raises questions related to microplanning: perhaps, we desire to merge together different pieces such as in the following example using \emph{backward writing} style.

\begin{quotation}
\ul{In some ways you need that discipline, don't you, to be a saver, to think, ``I won't get into debt''?.} [\textbf{T3}]

\noindent
\fbox{\textsf{Indeed}} 
\ul{If you want the economy to run smoothly, you have to incentivise certain type of behaviour.} [\textbf{T4}]
\fbox{\textsf{\sf , e.g.}} 
\ul{in South Korea, in terms of how South Korean grew, it did incentivise saving, at certain times, by certain economic policies} [\textbf{T5}]
\end{quotation}

\noindent
[\textbf{T4}] and [\textbf{T5}] are merged by a comma and \fbox{\sf e.g.}. In this way, we highlight the \emph{Evidence} relation and identify the connection as an argument from example instead of forming two independent sentences, which is an aspect traditionally related to microplanning. 

Finally, if we have knowledge that an inference rule \fulfils\ an argumentation scheme, then we can also map the elements of an argument into such a scheme. For instance, argument $\argb$ is classified as an argument by established rule, that is represented by the following scheme:

\noindent
\begin{tabu}{l l X[1,l]}
Major Premise: & & If carrying out types of actions including the state of affairs A is the established rule for x, then (unless the case is an exception), x must carry out A.\\
Minor Premise: & & Carrying out types of actions including state of affairs A is the
established rule for a.\\
Conclusion: & & Therefore a must carry out A.
\end{tabu}

It is interesting to note that the minor premise is left implicit in the formalisation of our example, and it might be an element that should be highlighted to the user. How to report an implicit premise may depend on the communicative goal. For example, if the system is intended for users to improve and make their arguments more explicit we may generate the text:


\begin{quotation}
    \ul{In some ways you need that discipline, don't you, to be a saver, to think, ``I won't get into debt''?.} [\textbf{T3}]
    \fbox{\textsf{Indeed}} 
    \ul{If you want the economy to run smoothly, you have to incentivise certain type of behaviour.} [\textbf{T4}]
    
    \noindent
    \fbox{\sf although we have no evidence that this is the established rule.} 
\end{quotation}

In this case, the fact that there is a lack of the minor premise is added to the generated text. On the other hand, if the system is intended to report an analysis of a conversation, we need to take into account that the premise is implicit as it may be already known by the participants. Hence,  the system could assume that the premise holds, and generate an explicit sentence \fbox{\sf and we assume that this is the established rule.} Further research is, however, needed in understanding how to generate text in case of unstated premises (e.g., in case of enthymemes\cite{waltonetal2008,Black2012}). With similar considerations,  we could also include critical questions that have either already been answered, or that, if answered, could strengthen the argument.

\subsection{Presenting an Argument Network}
\label{sec:presentingnetwork}

One of the strengths of formal argumentation is its ability to handle conflicts. In the previous section we focused on how to represent a single argument. However, even in our small running example, we can note that (1) there are more than one argument; and (2) there are conflicts among them (Figures \ref{fig:dundee1724} and \ref{fig:ex}).

The simplest strategy for representing an argument network is just to enumerate all the arguments and to list the conflicts among them, eventually linking that with critical questions if the information is available (e.g. \fbox{\sf This counterargument answers} \fbox{\sf the critical question that states \ldots}).

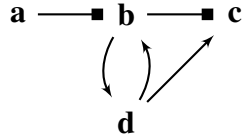
\begin{figure}[t]
  \centering
      \begin{tikzpicture}[->,>=stealth',shorten >=1pt,auto,node
      distance=1.2cm, thick,main node/.style={draw=none,fill=none},scale=1.2, every node/.style={scale=1.2}]
      
      \node[main node] (1) {$\arga$};
      \node[main node] (2) [right of=1] {$\argb$};
      \node[main node] (3) [below of=2] {$\argd$};
      \node[main node] (4) [right of=2] {$\argc$};

      \tikzset{edge/.style = {->,> = latex'}} 
      
      \draw[edge] (2) to[bend right] (3);
      \draw[edge] (3) to[bend right] (2);
      \draw[edge] (3) to (4);
      \draw [-{Square[]}]   (1) to (2);
        \draw [-{Square[]}]   (2) to (4);
    \end{tikzpicture}

  \caption{The \AFname{} $\AFsymbol{m}$ for the argument network of Figure \ref{fig:dundee1724} interpreted using deductive argumentation in a bipolar-inspired framework, where the squared arrows represents generic dependencies based on inferences.}
  \label{fig:exbipolar}
\end{figure}

However, we argue that there is a better approach, motivated by the need to satisfy Grice's maxim of relation and relevance \cite{Grice1975}; i.e. that in a conversation one needs to be relevant and say things that are pertinent to the discussion. Let us consider Figure \ref{fig:exbipolar} that depicts the $\AFsymbol{m}$ argumentation framework (cf. Fig. \ref{fig:ex}) annotated with a second, bipolar-inspired \cite{amgoudetal2008},\footnote{By \emph{bipolar-inspired} we intend that such a relation does not represent conflicts. At the same time, however, we are not in the position to argue that it would be a relationship of \emph{support}.} binary relation between arguments. Such a relation is grounded in the original argument network, Figure \ref{fig:dundee1724}.

Once such an annotated graph is obtained, a line of reasoning annotated together (e.g.\ \tuple{\arga, \argb, \argc} in Figure \ref{fig:exbipolar}) could be merged together in order to provide a single approximate argument. Once all the lines of reasoning annotated together are identified and merged, then we need a strategy to order their presentation; e.g.\ ordering by length of each line of reasoning versus ordering by the number of attacks received. We also need appropriate signals for \emph{antithesis} relations need to be used. For instance:

\begin{quotation}
    \ul{if you want the moral hazard, instead of kind of just going on about the bankers, is there not a danger that if we just said we'd write off debt, that it actually isn't very helpful for our side, for ordinary people, to actually have that?} [\textbf{T1}]
    \fbox{\sf Indeed}
    \ul{There's no discipline there} [\textbf{T2}] 
    \fbox{\sf and}
    \ul{In some ways you need that discipline, don't you, to be a saver, to think, ``I won't get into debt''?.} [\textbf{T3}]
    \fbox{\textsf{Indeed}} 
    \ul{If you want the economy to run smoothly, you have to incentivise certain type of behaviour.} [\textbf{T4}]
    \fbox{\textsf{\sf , e.g.}} 
    \ul{in South Korea, in terms of how South Korean grew, it did incentivise saving, at certain times, by certain economic policies} [\textbf{T5}]
    \fbox{\sf However,}
    \ul{people don't realise, or only half realise, is the fact that we have actually written off massive amounts of debt} [\textbf{T6}]
        \fbox{\sf and}
    \ul{But it certainly isn't the debts of the people who most need it in society} [\textbf{T7}].
\end{quotation}

\subsection{Explaining the Acceptability Status of a Single Argument or an Approximate Argument}
\label{sec:explainacceptability}

Two of the traditional decision problems in abstract argumentation \cite{dw:2009} are checking whether an argument is credulously or skeptically accepted (or not). An interested user might select one of the arguments (or one of the propositions in the argument network) and ask whether or not it is credulously or skeptically accepted.

\begin{figure}[t]
  \centering
      \begin{tikzpicture}[-,>=stealth',shorten >=1pt,auto,node
      distance=1.2cm, thick,main node/.style={draw=none,fill=none},scale=1.2, every node/.style={scale=1.2}]
      
      \node[main node] (1) {$\argc$};
      \node[main node] (2) [below of=1] {$\argd$};
      \node[main node] (3) [below of=2] {$\argb$};
      \node[main node] (4) [left of=1] {PRO};
      \node[main node] (5) [left of=2] {CON};
      \node[main node] (6) [left of=3] {PRO};

      \tikzset{edge/.style = {-,> = latex'}} 
      
      \draw[edge] (1) to (2);
      \draw[edge] (2) to (3);
    \end{tikzpicture}

  \caption{Dispute tree to prove that \argc\ is credulously accepted w.r.t. preferred semantics in the \AFname{} $\AFsymbol{m}$ for the argument network of Figure \ref{fig:dundee1724} interpreted using deductive argumentation.}
  \label{fig:proof}
\end{figure}
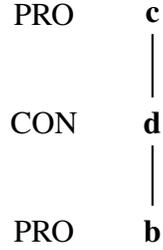

To answer such a query, we can exploit research in proof theory and argument games \cite{Modgil2009}. For instance, to prove that \argc\ (in Figure \ref{fig:ex}) is credulously accepted according to preferred semantics, we can (1) compute the extensions; and (2) compute the dispute tree \cite{Modgil2009} needed to prove it. The dispute tree is depicted in Figure \ref{fig:proof} which suggests that only the argument network comprising \argc, \argd, and \argb\ needs to be considered.

\subsection{Explaining Extensions}

Explaining multiple extensions is analogous to present different argumentation \linebreak \mbox{(sub-)networks} that are linked together by attacks. Therefore, the solution for this communication goal builds upon the procedures envisaged in Sections \ref{sec:presentingnetwork} and \ref{sec:explainacceptability}.

It is worth mentioning that a possible way to identify the attack connections between those extensions is to consider a set of arguments belonging to a set of issues $\aset \in \setgenext{\mbox{issues}}{\anAFsymbol}$ such that $|\aset| = |\setgenext{\sigma}{\anAFsymbol}|$ for the chosen $\sigma$ semantics.\footnote{Assuming that the semantics can be related to the notion of a complete labelling (Def. \ref{def:issues}).} Those arguments could nicely provide foci of attention: for instance, $\set{\argb, \argd} \in \setgenext{\mbox{issues}}{\AFsymbol{m}}$. Therefore, the text presented in Section \ref{sec:presentingnetwork} could be adapted in a straightforward manner to communicate the existence of two preferred extensions that gravitate around the issue $\set{\argb, \argd}$.

\section{Conclusion}
\label{sec:conclusion}

In this paper have we provided a blueprint for a complex set of research questions that arise when considering how to generate natural language representations of formal argumentation structures.

First of all, as already noted at the beginning of Section \ref{sec:nlgarg}, we assume neither a model of the user, nor pre-existing contexts that need to be referenced in the piece of text generated. These elements will need to be addressed in future research.

Moreover some pieces of generated text sound quite awkward. To address such an issue, each piece of information inserted in an argument network should represent a single \emph{text-agnostic normal form} proposition. In this way, we could compose them in a way in a meaningful way. This is clearly a constraint that might be unnatural if an untrained user tries to generate an argument network, and this raises an interesting research question concerning how to enable untrained users to formalise their lines of reasoning.

It is worth noting that the research questions highlighted in this paper have been grounded on a piece of argumentation formalised from a discussion between humans. We claim neither that our list is exhaustive nor that it is applicable to all possible argument networks. For instance, providing a natural language interface to an argument network built by an expert might lead to different, unforeseen, communicative goals.  For instance, related research enabling the scrutiny of autonomous systems by allowing agents to generate plans through argument and dialogue \cite{Tintarev2013} had the specific communication goal of justifying the purpose of each step within a plan. 

Moreover we support the effort of extending AIF to include additional information: for instance, the AIFdb \cite{Lawrence2012} project extends the AIF model to include schemes: of rephrasing; of locution describing communicative intentions which speakers use to introduce propositional contents; and  of interaction or protocol describing relations between locutions. Dealing with those pieces of information raise further research questions. Indeed, as we saw in Section \ref{sec:simplelogic}, arguments \arga, \argb, and \argc\ form a conflict-free set, and in Section \ref{sec:nlgarg} we often considered them altogether in an approximate argument. However, while argument \argc\ was put forward by Claire Fox in the original dialogue (Section \ref{sec:running-example}), arguments \arga\ and \argb\ belong to Nick Dearden. The question here is how to include the sources of those arguments in the generated natural language text. In the use of argumentation to support reasoning and intelligence analysis \cite{Toniolo2015}, sources are also important, but this is only one aspect of the provenance of an argument that could, in general, be considered.

Finally, we will investigate the potential of applying NLG to existing systems using formal argumentation in real-world applications, such as CISpaces \cite{Toniolo2015} --- a system for collaborative intelligence analysis --- and ArgMed \cite{Hunter2012,Williams2015} --- a system for reasoning about the results of clinical trials.

\bibliographystyle{apalike}
\bibliography{biblio}

\end{document}